\documentclass[letterpaper, 10 pt, conference]{ieeeconf}  

\usepackage{amsmath}
\usepackage{graphicx}
\usepackage{bm}
\usepackage{amssymb}
\usepackage{subcaption}
\usepackage{algorithm}
\usepackage{algorithmic}
\usepackage{siunitx}
\usepackage{multirow}
\usepackage{array}
\usepackage[colorlinks,bookmarksopen,bookmarksnumbered,citecolor=blue,urlcolor=blue]{hyperref}
\usepackage{float}
\usepackage{tikz}
\usepackage{atbegshi}

\IEEEoverridecommandlockouts                              

\title{\LARGE \bf
Dynamic Modeling and MPC for Locomotion of Tendon-Driven Soft Quadruped
}

\author{Saumya Karan$^{1}$, Neerav Maram$^{2}$, Suraj Borate$^{1}$ and Madhu Vadali$^{1}$
\thanks{$^{1}$Saumya Karan is a PhD scholar in the Department of Electrical Engineering, IIT Gandhinagar, Gujarat, India.
        {\tt\small karansaumya@iitgn.ac.in}}%
\thanks{$^{2}$Neerav Maram is a BTech scholar in the Department of Electronics and Computer Engineering, Sreenidhi Institute of Science and Technology, Telangana, India.
        {\tt\small 22311a1901@ecm.sreenidhi.edu.in}}%
\thanks{$^{1}$Suraj Borate is a PhD scholar in the Department of Mechanical Engineering, IIT Gandhinagar, Gujarat, India.
        {\tt\small surajb@iitgn.ac.in}}%
\thanks{$^{1}$Madhu Vadali is an Associate Professor in the Department of Mechanical Engineering, IIT Gandhinagar, Gujarat, India.
        {\tt\small madhu.vadali@iitgn.ac.in}}%
}

\begin{document}
\AtBeginShipout{%
  \AtBeginShipoutUpperLeft{%
    \begin{tikzpicture}[remember picture,overlay]
      \node[anchor=north east] at ([xshift=-5mm,yshift=-5mm]current page.north east)
      {\small\bfseries Preprint};
    \end{tikzpicture}
  }
}

\maketitle

\thispagestyle{empty}
\pagestyle{empty}

\begin{abstract}
SLOT (Soft Legged Omnidirectional Tetrapod), a tendon-driven soft quadruped robot with 3D-printed TPU legs, is presented to study physics-informed modeling and control of compliant legged locomotion using only four actuators. Each leg is modeled as a deformable continuum using discrete Cosserat rod theory, enabling the capture of large bending deformations, distributed elasticity, tendon actuation, and ground contact interactions. A modular whole-body modeling framework is introduced, in which compliant leg dynamics are represented through physically consistent reaction forces applied to a rigid torso, providing a scalable interface between continuum soft limbs and rigid-body locomotion dynamics. This formulation allows efficient whole-body simulation and real-time control without sacrificing physical fidelity. The proposed model is embedded into a convex model predictive control framework that optimizes ground reaction forces over a 0.495 s prediction horizon and maps them to tendon actuation through a physics-informed force–angle relationship. The resulting controller achieves asymptotic stability under diverse perturbations. The framework is experimentally validated on a physical prototype during crawling and walking gaits, achieving high accuracy with less than 5 mm RMSE in center of mass trajectories. These results demonstrate a generalizable approach for integrating continuum soft legs into model-based locomotion control, advancing scalable and reusable modeling and control methods for soft quadruped robots. Codes and design available at \href{https://github.com/firstlastnamepaper-beep/SLOT_Github}{SLOT\_Github}.
\end{abstract}


\section{INTRODUCTION}

Soft quadruped robots offer unique advantages that make them particularly helpful in various applications. Their inherent flexibility and compliance allow for safer interactions with humans and the environment, making them well-suited for tasks in complex or sensitive settings \cite{c1}.

However, the accurate modeling and real-time control of such soft robotic systems remain challenging due to their nonlinear, large deformation behaviors and dynamic interactions with varied terrains \cite{c2}. Traditional approaches to robot modeling have often relied on simplified representations, particularly for fixed-end manipulators, leading to limitations in position and velocity estimation accuracy. Conventional modeling approaches, such as rigid-body dynamics \cite{c3}, constant-curvature assumptions \cite{c4}, and lumped-mass-spring models \cite{c5} struggle with complex, deformable structures like soft robotic limbs. While computationally efficient, they fail to capture the nonlinear and large deformations of soft materials, leading to discrepancies between simulated and actual behaviors, particularly during foot-ground impacts, tendon-driven actuation, and load-bearing phases \cite{c6}.

The finite element method (FEM) is a standard tool for modeling deformable structures due to its physical accuracy. However, its high computational cost, especially the repeated evaluation of stiffness matrices for nonlinear deformations, limits its suitability for real-time control in mobile robots \cite{c7}. While FEM yields detailed insights into stress and material behavior, its implementation in control loops remains impractical \cite{c8}. Particle-based models like mass-spring systems and position-based dynamics offer better computational efficiency by representing structures as particle networks. Though intuitive and faster, these models lack the physical realism needed to capture complex material responses and volumetric deformation — critical for soft robotic legs in locomotion tasks. The boundary element method (BEM) reduces computation by modeling only surface deformations, but it assumes homogeneous interior properties, restricting its use in robots with complex, layered soft structures \cite{c10}.

Constitutive models aim to capture the physical properties of materials with high fidelity, providing a foundation for realistic simulations. Examples include linear elastic models like Hookean solids for small deformations \cite{c11}, and hyperelastic models such as Neo-Hookean \cite{c12}, Mooney-Rivlin, or Ogden \cite{c14} for large, nonlinear strains common in soft robotics. Viscoelastic models like Kelvin-Voigt also account for time-dependent behavior. While these models offer excellent physical accuracy, the challenge lies in balancing detailed material representation with the need for real-time performance in mobile robot applications.

Machine learning has become a powerful tool for robot modeling and control. Supervised learning can capture complex input-output mappings but demands large labeled datasets \cite{c16}. Reinforcement learning learns through trial and error but requires extensive training \cite{c17}. Imitation learning offers a compromise by learning from demonstrations, though it may struggle to generalize beyond them.

The Cosserat rod theory offers a high-fidelity framework for modeling slender, deformable structures in soft robotics, addressing key limitations of traditional approaches like rigid-link and constant-curvature models \cite{c19}. Unlike these simplified methods, Cosserat rods provide a continuum mechanics foundation capable of capturing complex deformation modes such as bending, twisting, and stretching—features essential for soft limbs undergoing dynamic gait cycles \cite{c20}. This is particularly relevant for mobile quadrupeds, where real-time interaction with varying terrains demands accurate modeling of compliant leg mechanics to ensure stable and adaptive locomotion.

Although a few soft and quasi-soft quadruped robots have been reported in the literature, most existing platforms primarily focus on demonstrating novel actuation mechanisms or locomotion capabilities. Representative works employ pneumatic or inflatable soft actuators \cite{c29}, distributed soft or tendon-driven actuation schemes \cite{c30}, or learning-based gait generation with simplified internal models \cite{c31}. While effective for showcasing compliant locomotion, these approaches typically do not emphasize physics-based leg modeling integrated with real-time, model-predictive whole-body control. In contrast, tendon-driven soft quadrupeds with a minimal actuator count remain relatively underexplored from a first-principles dynamics and control perspective. The present work addresses this gap by combining Cosserat rod–based modeling of compliant TPU legs with a decoupled whole-body formulation and real-time convex MPC on embedded hardware, enabling experimentally validated, stable locomotion.

Motivated by this gap, in this work we present SLOT, a tendon-driven soft quadruped robot with 3D-printed TPU legs, and develop a physics-informed modeling and control framework based on discrete Cosserat rod theory and real-time model predictive control to achieve stable and accurate locomotion using only four actuators. Cosserat rod theory is employed to model the compliant TPU legs, extending its application beyond traditional fixed-end continuum manipulators to a mobile quadruped setting. The nonlinear viscoelastic behavior of TPU poses modeling challenges that require distributed force and moment application along the leg - something that Cosserat rods naturally handle. This enables a precise simulation of ground contact, load distribution, and coupled deformations critical for gait transitions such as walking and crawling. Leveraging this approach in a quadruped setting introduces new challenges, including the need to model multiple legs simultaneously and capture the dynamic coupling between leg deformation and torso motion.

Given the computational complexity of simultaneously modeling torso and multiple compliant legs, we adopt a decoupled modeling approach inspired by the MIT Cheetah robot \cite{c21}. This modular approach simulates each leg's dynamics independently, aggregating resulting reaction forces and torques to update the torso dynamics efficiently. Such decoupling significantly reduces computational demands, making it feasible for real-time model predictive control (MPC).

MPC is crucial for adaptive real-time multivariable control \cite{c22}, enabling trajectory optimization and stable locomotion. SLOT, the soft quadruped robot under study, comprises a rigid cuboid torso supported by four 3D-printed flexible TPU legs, each tendon-actuated by a single servo motor for compliant, omnidirectional locomotion. This paper contributes by combining Cosserat rod-based decoupled modeling of SLOT with MPC, providing robust and computationally efficient control strategies tailored explicitly for soft quadruped robots. Through experimental validation, this research demonstrates improved accuracy in predicting soft limb deformation and robot locomotion dynamics, representing a significant advancement in physics-informed design for soft robotics.

This paper contributes a physics-informed modeling and control framework that enables real-time, whole-body locomotion control for tendon-driven soft quadruped robots. The key contributions are as follows:
\begin{enumerate}
    \item A discrete Cosserat rod model is formulated for tendon-driven soft legs, capturing large deformation, distributed elasticity, and ground contact while remaining suitable for real-time locomotion control.

    \item A modular whole-body modeling principle is introduced, in which continuum leg dynamics are interfaced with rigid-body torso motion through physically consistent reaction forces, enabling scalable integration of soft limbs into legged locomotion frameworks.

    \item The resulting soft–rigid hybrid dynamics are embedded into a convex model predictive control (MPC) formulation that optimizes ground reaction forces and maps them to tendon actuation via a physics-informed force–angle relationship.

    \item A full hardware-in-the-loop locomotion pipeline is implemented using a ROS2-based architecture, integrating visual–inertial state estimation, predictive control, and low-level tendon actuation on embedded hardware.

    \item The proposed framework is experimentally validated on a tendon-driven soft quadruped robot, demonstrating accurate deformation prediction, stable multi-gait locomotion, and robustness to structured perturbations.
\end{enumerate}

\section{SYSTEM OVERVIEW}
\begin{figure*}[h!]
\centering
\includegraphics[width=0.6\textwidth]{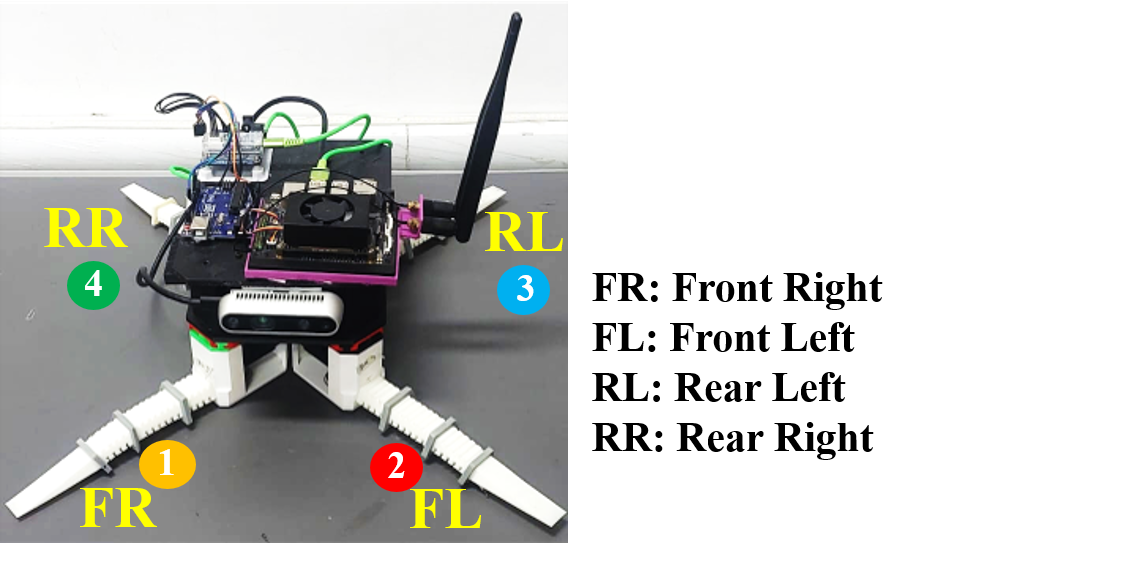} 
\caption{SLOT robot hardware}
\label{fig:TetraFlex_hardware}
\end{figure*}

From our previous work, SLOT \cite{c23} is a novel tendon-driven, soft quadruped that employs only four actuators and demonstrates diverse capabilities: sit/stand postures, roll and pitch maneuvers, turning, lateral, longitudinal, as well as omnidirectional locomotion. SLOT has a total mass of 2.16 kg (2.0 kg for the torso and 0.04 kg per leg), with the rigid cuboid torso measuring 125.5 mm × 85.5 mm × 34 mm. Using only four Dynamixel XL430 servo motors (1.4 Nm torque) and flexible 3D printed TPU legs, SLOT achieves motion in six degrees of freedom while maintaining compliance. The robot, shown in Fig. \ref{fig:TetraFlex_hardware}, operates autonomously with a fully on-board stack running on an NVIDIA Jetson Xavier (8 GB RAM, 6-core ARM CPU, 384-core NVIDIA Volta GPU), requiring no off-board computation or teleoperation. Low-level PID controllers regulate body height, while an Intel RealSense D435 (1280×720 RGB-D at 30 Hz) supplies depth for visual SLAM and orientation estimation; a wireless link is used only for telemetry and logging. SLOT employs handcrafted gaits for crawling, walking, turning, and omnidirectional movement, with a unique diagonal leg placement enabling efficient maneuverability. The robot demonstrates a variety of motions, including crawling under obstacles, climbing inclines up to 20°, and maintaining stability under external disturbances \cite{c23}. Open-source and highly adaptable, SLOT has the potential for real-world applications such as exploration, search-and-rescue, and bio-inspired robotics.

\section{SINGLE LEG DYNAMIC MODEL}

The design and diverse motions of SLOT, necessitate a robust model to capture the complex dynamics of its soft legs.  A Cosserat rod framework is employed to simulate each leg, capturing its large, nonlinear deformations with high fidelity. The model incorporates internal elasticity, tendon-induced actuation, ground interaction, and damping effects. This modeling approach is essential not only to simulate realistic deformations but also to narrow down the feasible control space, enabling efficient optimization and stable locomotion.

\subsection{Assumptions}
Each leg, made of TPU, is modeled as a planar Cosserat rod constrained to deform in the XZ plane. This planar assumption is adopted to enable real-time simulation and control and reflects the observation that the dominant leg deformation arises from bending in the local sagittal plane, with out-of-plane bending and twist observed to be negligible for the tested gaits. It is discretized into $N = 31$ nodes along a total length of $L = 0.19$ m, with a linearly tapered thickness from 13.5 mm at the node fixed to the torso to 3.5 mm at the free tip to approximate realistic leg geometry as shown in Fig. \ref{fig:heighttimelapse}. The choice of discretizing each leg into 31 nodes is motivated by a trade-off between numerical accuracy and real-time computational cost. The TPU material is assumed to be homogeneous, isotropic, and linearly elastic with a Young's modulus of $E = 1 \times 10^7$ Pa and density $\rho = 1200\,\text{kg/m}^3$. This linear elastic approximation is adopted to capture dominant deformation trends relevant to locomotion control, while higher-order viscoelastic effects are compensated through feedback control and MPC receding-horizon updates. Actuation is achieved via a tendon routed through a fixed virtual pulley positioned at a constant angle of 168° with respect to horizontal axis. Ground interaction is modeled through a penalty-based spring-damper system \cite{c25} that produces only vertical contact forces. Additionally, Coulomb friction is applied to the leg to resist sliding during ground contact. Table \ref{tab:simulation_params} shows all the simulation and material parameters for the single leg model.

\begin{table}[h!]
\centering
\caption{Simulation and Material Parameters}
\begin{tabular}{|l|c|l|}
\hline
\textbf{Parameter} & \textbf{Value} & \textbf{Description} \\
\hline
$L$ & 0.190 m & Total leg length \\
$N$ & 31 & Number of nodes \\
$dx$ & $L / (N-1)$ & Segment length \\
$w$ & 0.02 m & Constant leg width \\
$h_{\text{base}}$ & 0.0135 m & Thickness at fixed base \\
$h_{\text{tip}}$ & 0.0035 m & Thickness at tip \\
$E$ & $1 \times 10^7$ Pa & Young's modulus (TPU) \\
$\rho$ & 1200 kg/m$^3$ & Density of TPU \\
$g$ & -9.81 m/s$^2$ & Gravitational acceleration \\
$\mu$ & 0.1 & Coulomb friction coefficient \\
$c$ & 0.1 & Linear damping coefficient \\
$k_{\text{contact}}$ & $1 \times 10^5$ N/m & Ground stiffness \\
$d_{\text{contact}}$ & 10 Ns/m & Ground damping \\
$k_r$ & 800 N/m & Restoring spring stiffness \\
$\theta_{\text{pulley}}$ & $168^\circ$ & Pulley angle relative to base \\
$R_0$ & 0.03 m & Pulley radius offset \\
$\Delta t$ & $1 \times 10^{-4}$ s & Time step \\
$T_{\text{max}}$ & 1.5 Nm & Max tendon tension \\
Angle & 116.28° & Max servo pulling angle \\
\hline
\end{tabular}

\label{tab:simulation_params}
\end{table}

\subsection{From Continuous to Discrete Cosserat Rod Formulation}

The Cosserat rod theory provides a continuum-mechanics description of slender deformable bodies, in which the rod configuration is parameterized by a spatial variable $s \in [0,L]$ and time $t$. At each material cross-section, the state is described by a position vector $\mathbf{r}(s,t) \in \mathbb{R}^3$ and an orientation $\mathbf{R}(s,t) \in SO(3)$. The governing equations follow from the balance of linear and angular momentum and result in a coupled set of nonlinear partial differential equations (PDEs) along the rod length.

In this work, a \emph{space-discretized Cosserat rod model} is employed, in which the continuous rod is approximated by a finite number of rigid cross-sections connected through elastic constitutive relations. The arc-length domain is discretized into $N$ nodes with uniform spacing $\Delta s = L/(N-1)$. Each node $i$ represents a Cosserat cross-section with generalized coordinates
\begin{equation}
\mathbf{q}_i = [x_i,\; z_i,\; \theta_i]^{\mathsf{T}},
\end{equation}
corresponding to planar position and orientation.

Spatial discretization converts the continuous Cosserat rod PDEs into a finite-dimensional system of ordinary differential equations (ODEs), where internal elastic forces and moments arise from relative strains between adjacent nodes. Axial deformation is captured through stretch between neighboring nodes, while bending is modeled through changes in relative orientation, consistent with Cosserat rod constitutive laws. These internal forces and moments are computed locally at the segment level and assembled into nodal force and torque balances using Newton--Euler equations.

Time remains continuous in the modeling stage, and no temporal discretization is introduced in the formulation. The resulting system preserves the essential characteristics of Cosserat rod mechanics, including distributed elasticity, large deformation capability, and force--moment transmission, while enabling efficient numerical simulation and integration with whole-body dynamics and control. This approach follows standard discrete Cosserat rod formulations commonly used in computational rod mechanics and soft robotics literature.

The approach follows the methodology introduced by \cite{c26} and \cite{c27}, where bending and stretching energies are defined between adjacent segments, resulting in a finite-dimensional system of ordinary differential equations governing the nodal dynamics. Unlike reduced-order Cosserat models \cite{c19} based on piecewise-constant strain approximations, the present formulation directly computes internal forces and moments at the nodal level, preserving consistency with continuum Cosserat rod mechanics while enabling efficient numerical integration.

\subsection{Model Calibration Procedure}

Model parameters were calibrated prior to experimental validation using a physics-informed procedure. Geometric properties of each leg were taken directly from the CAD model used for fabrication. Material density and initial elastic modulus values for TPU were obtained from manufacturer specifications, with the Young’s modulus refined using quasi-static single-leg deflection tests under known tendon actuation.

Ground contact stiffness, damping, and friction coefficients were chosen based on standard penalty-based contact modeling practices commonly used in legged and soft robotic simulations. The selected values ensure numerically stable contact and were kept constant across all experiments. Linear damping and restoring stiffness terms were introduced primarily for numerical stabilization of the discretized rod dynamics and were chosen sufficiently small so as not to dominate the physical elastic response.

The mapping between servo pulling angle and vertical ground reaction force was obtained from independent single-leg experiments and fixed prior to whole-body locomotion trials. No model parameters were tuned using full-body locomotion data, ensuring an independent validation of the proposed model.

\subsection{Force–Moment Balance and Numerical Simulation}

The soft leg’s motion is governed by the balance laws of linear and angular momentum, ensuring equilibrium between internal and external forces. At each timestep, the following physical effects are computed:

\paragraph{Internal Forces and Moments}
For each segment between nodes $i$ and $i+1$, the axial strain is defined as:
\[
\varepsilon_i = \frac{l_i - \Delta s}{\Delta s}
\] 
where $l_i = \sqrt{(x_{i+1} - x_i)^2 + (z_{i+1} - z_i)^2}$ is the current length of segment and $\Delta s$ is the natural/undeformed length. The resulting internal axial force is: $f_{\text{axial},i} = E_i A_i \varepsilon_i$. This force is projected along the local orientation $\theta_i = \arctan2(z_{i+1} - z_i, x_{i+1} - x_i)$ and applied in opposite directions at the connected nodes in the local frame:
\[
\mathbf f^{\text{loc}}_{i}=\begin{bmatrix} +f_{\text{axial},i}\\[2pt] 0\end{bmatrix},\qquad
\mathbf f^{\text{loc}}_{i+1}=\begin{bmatrix} -f_{\text{axial},i}\\[2pt] 0\end{bmatrix}
\]

Transforming to the global $(x,z)$ frame, the forces are updated as,
\[
\begin{bmatrix}F_{x,i}\\ F_{z,i}\end{bmatrix}\!\leftarrow\!\begin{bmatrix}F_{x,i}\\ F_{z,i}\end{bmatrix}+\mathbf R_i^{\mathsf T}\,\mathbf f^{\text{loc}}_{i},\qquad
\begin{bmatrix}F_{x,i+1}\\ F_{z,i+1}\end{bmatrix}\!\leftarrow\!\begin{bmatrix}F_{x,i+1}\\ F_{z,i+1}\end{bmatrix}+\mathbf R_i^{\mathsf T}\,\mathbf f^{\text{loc}}_{i+1}
\]
where, $\mathbf R_i=\begin{bmatrix}\cos\theta_i & \sin\theta_i\\ -\sin\theta_i & \cos\theta_i\end{bmatrix}$

The change in orientation across segments, $\Delta\theta_i=\theta_{i+1}-\theta_i$, is used to capture bending. The bending moment is computed as $m_{\text{bend},i}=\frac{E_i I_i}{\Delta s}\,\Delta\theta_i$, and applied as internal torques $M_{y,i}\!\leftarrow\! M_{y,i}-m_{\text{bend},i}$, $M_{y,i+1}\!\leftarrow\! M_{y,i+1}+m_{\text{bend},i}$.

\noindent
Here, \( E_i \) denotes the Young's modulus, \( A_i \) is the cross-sectional area computed as \( w \cdot h_i \), and \( I_i \) is the second moment of area given by \( \frac{w h_i^3}{12} \), where \( h_i \) is the local rod thickness and \( w \) is the constant rod width. These parameters determine the rod’s elastic and bending response at each segment.

At every timestep, these internal forces ($F_x$, $F_z$) and moments ($M_y$) are summed with other physical effects including tendon forces, damping, contact, and friction, to compute nodal accelerations using Newton-Euler equations.

\paragraph{Tendon Tension}
A time-varying tendon force $T(t)$ is applied along \(\frac{2}{3}\)rd of the leg length. The tendon force follows a time-dependent profile comprising three phases: a ramp-up phase using a sine function, a short hold phase where the maximum tension is maintained, and a decay phase modeled by a cosine function that gradually reduces the tension back to zero, emulating servo torque behavior.

\paragraph{Ground Contact and Friction}
Each node exerts a ground contact force when penetrating below $z=0$,
\begin{equation}
F_{\text{contact}, i} =
\begin{cases}
k_{\text{contact}} \cdot \delta_i + d_{\text{contact}} \cdot \dot{\delta}_i & \text{if } z_i < 0 \\
0 & \text{otherwise}
\end{cases}
\end{equation}
with penetration depth $\delta_i = -z_i$ and normal velocity $\dot{\delta}_i = -v_{z,i}$.
Assuming Coulomb friction, the force is applied only when the leg nodes are in contact with the ground:
\begin{equation}
F_{\text{friction}} =
\begin{cases}
-\mu (m + \frac{M}{4}) g \cdot \text{sign}(v_x) & \text{if } z \leq 0 \\
0 & \text{otherwise}
\end{cases}
\end{equation}
where $\mu$ is the coefficient of friction, $m$ is the mass of the leg and $M$ is the mass of the robot

\paragraph{Restoring Force to Initial Configuration}

Each node is pulled back to its initial position using a virtual spring:
\[
F_{x, \text{restore}, i} = -k_r (x_i - x_{i,0}), \text{ and } F_{z, \text{restore}, i} = -k_r (z_i - z_{i,0})
\]
where $k_r$ is the restoring stiffness.

\paragraph{Damping Force}

Linear damping is applied to all degrees of freedom:
\[
F_{x, \text{damp}, i} = -c \cdot v_{x,i}, F_{z, \text{damp}, i} = -c \cdot v_{z,i}, \text{ and } M_{y, \text{damp}, i} = -c \cdot \omega_i
\]

\paragraph{Cosserat Rod Dynamics Formulation}

The system dynamics for a Cosserat rod can be written in the standard robot dynamics form:

\[
\boxed{
\bm{M(q)} \ddot{\bm{q}} + \bm{C(q, \dot{q})} \dot{\bm{q}} + \bm{g(q)} = \bm{T(q, t)}}
\]
\noindent
where \(\bm{q} = [x_0, z_0, \theta_0, \ldots, x_N, z_N, \theta_N]^\top\) is the generalized coordinate vector; \(\dot{\bm{q}}\) and \(\ddot{\bm{q}}\) are the velocity vector and acceleration vectors, respectively. $\mathbf{M}(q)$ is the mass matrix, $\bm{C(q, \dot{q}) \dot{q}}$ represents the damping and restoring force term, $\mathbf{g(q)}$ is the gravity vector and $\bm{T(q, t)}$ is the net applied force vector. These are given by,

\[
\mathbf{M}(q)=
\mathrm{diag}\!\left(m_0, m_0, I_0, m_1, m_1, I_1, \ldots, m_N, m_N, I_N\right)
\]
\[
\bm{C(q, \dot{q}) \dot{q}} = 
\begin{bmatrix}
c \dot{x}_0 + k(x_0 - x_{0,\text{init}}) \\
c \dot{z}_0 + k(z_0 - z_{0,\text{init}}) \\
c \omega_0 \\
\vdots \\
c \dot{x}_N + k(x_N - x_{N,\text{init}}) \\
c \dot{z}_N + k(z_N - z_{N,\text{init}}) \\
c \omega_N
\end{bmatrix}
\]

\[
\mathbf{g(q)} =
\begin{bmatrix}
0 & -m_0 g & 0 & \cdots & 0 & -m_N g & 0
\end{bmatrix}^{\mathsf{T}}
\]


\[
\bm{T(q, t)} = 
\bm{F}_{\text{internal}} +
\bm{F}_{\text{friction}} +
\bm{F}_{\text{contact}} +
\bm{F}_{\text{tendon}}
\]
\noindent
where \( m_i \) is the mass of the \( i \)-th node and \( I_i = \frac{1}{2} \rho A_i dx^2 \) is the moment of inertia of the \( i \)-th segment. \( c \) is the linear damping coefficient, and \( k \) is the restoring stiffness  to maintain structural integrity.
\subsection{Node-Level Dynamic Equations}

For each node \( i \in \{0, \ldots, N\} \), the dynamic equations are,
\[
\bm{M}_i
\begin{bmatrix}
\ddot{x}_i \\
\ddot{z}_i \\
\dot{\omega}_i
\end{bmatrix}
+
\begin{bmatrix}
c \dot{x}_i + k(x_i - x_{i,\text{init}}) \\
c \dot{z}_i + k(z_i - z_{i,\text{init}}) \\
c \omega_i
\end{bmatrix}
+
\begin{bmatrix}
0 \\
-m_i g \\
0
\end{bmatrix}
=
\begin{bmatrix}
F_{x,i}^{\text{total}} \\
F_{z,i}^{\text{total}} \\
M_{y,i}^{\text{total}}
\end{bmatrix}
\]

\begin{figure*}[h!]
\centering
\includegraphics[width=0.5\textwidth]{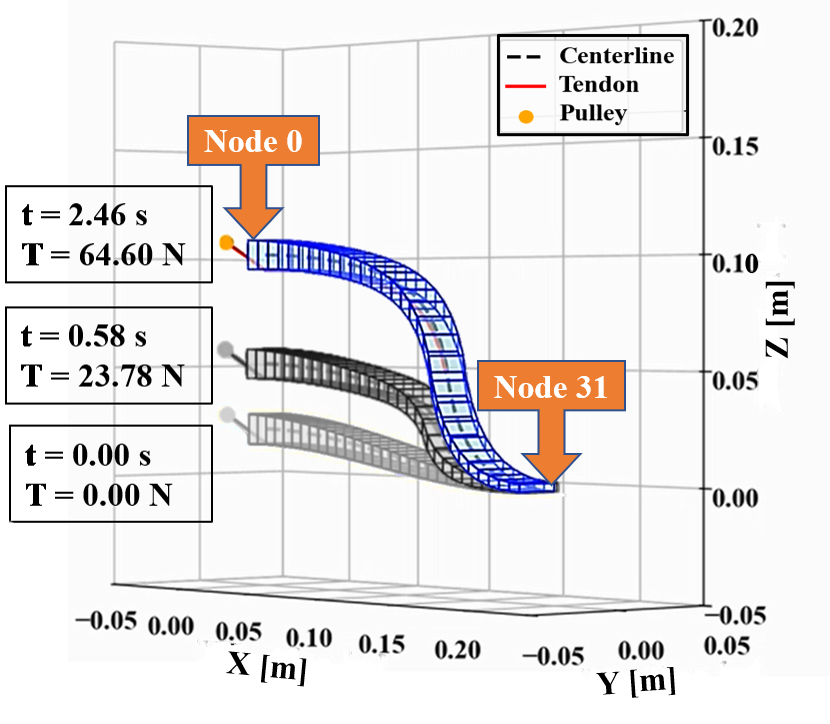} 
\caption{Single-leg deformation under tendon actuation.}
\label{fig:heighttimelapse}
\end{figure*}

This formulation represents the Newton-Euler equations for each node, accounting for linear and angular accelerations. The left-hand side aggregates inertial, damping, restoring, and gravitational effects, while the right-hand side represents the total forces and moments acting on the node due to elasticity, contact, friction, and tendon tension. Fig. \ref{fig:heighttimelapse} shows single leg deformation under tendon actuation.

The velocities and positions are then updated according to the explicit Euler integration. A sufficiently small time step ($\Delta t = 10^{-4}$~s) is used to maintain numerical stability for the chosen timestep. This explicit integration scheme enables efficient forward simulation of the discrete Cosserat rod dynamics while maintaining consistency with the underlying Newton--Euler formulation.

\section{SLOT WHOLE-BODY MODELING}
Building on the Cosserat rod formulation for individual legs, this section extends the model to the full SLOT system, integrating leg dynamics with torso motion.

\subsection{Assumptions}
\begin{figure*}[h!]
\centering
\includegraphics[trim=0 0 0 65,clip,width=\linewidth]{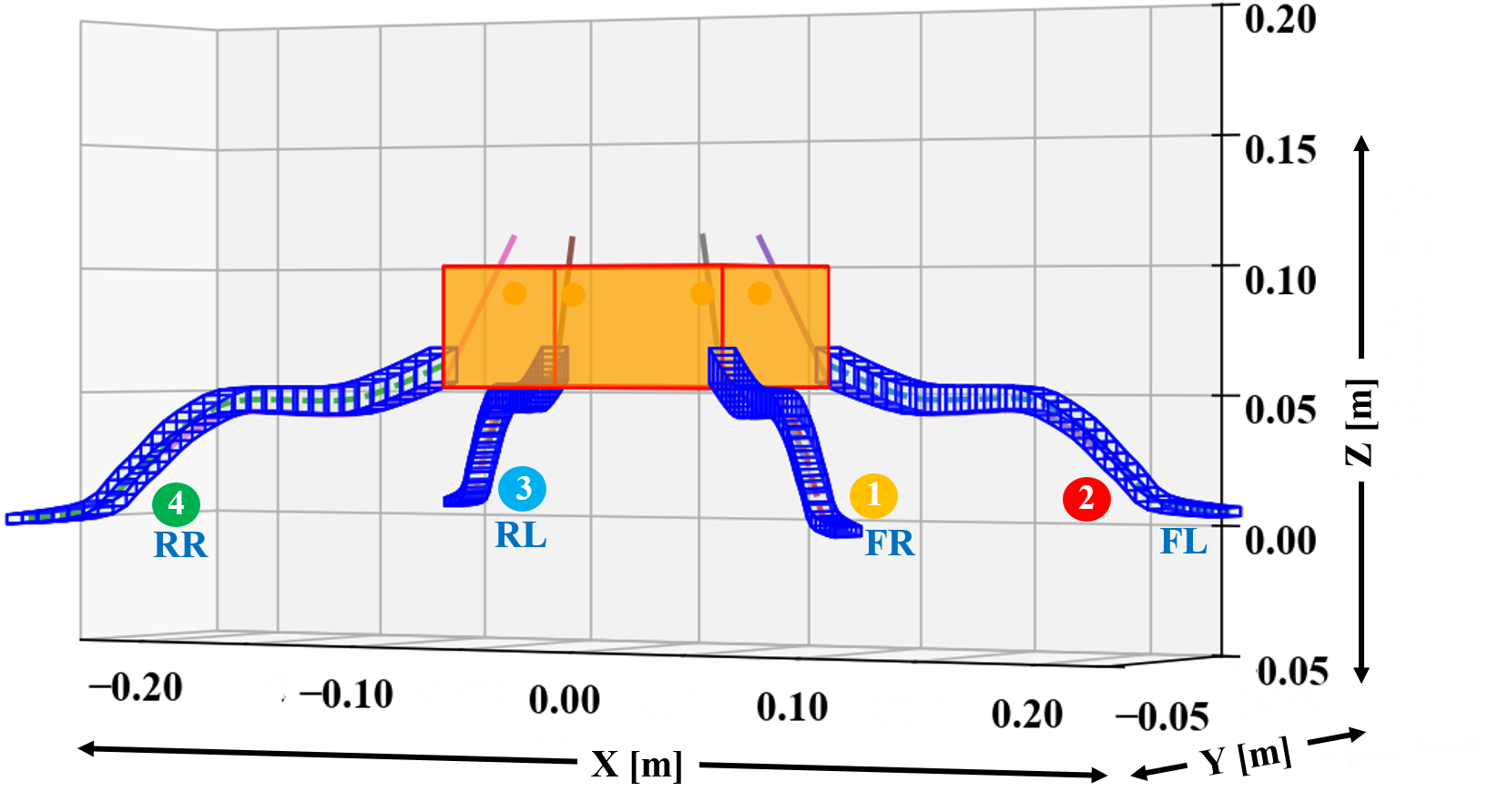} 
\caption{Force application from four Cosserat rod legs onto the rigid cuboid torso.}
\label{fig:full_slot}
\end{figure*}

The torso is modeled as a rigid cuboid with six degrees of freedom. Fig. \ref{fig:full_slot} shows the force application from four Cosserat rod legs onto the rigid cuboid torso. Each of the four legs is modeled independently using a planar Cosserat rod formulation and is rigidly attached to the torso at predefined corner locations (FL, BR, BL, FR). Ground contact forces are applied only at the leg tips. The interaction between each leg and the torso is captured through reaction forces and torques transmitted at the attachment points. Tendon tension profiles are defined separately for each leg and evolve over time according to prescribed actuation schedules. Both the torso and legs are subjected to gravitational, damping, and restoring forces throughout the simulation.

\subsection{Dynamics and Modular Coupling}
The full-body system involves dynamic interactions between the rigid torso and the four compliant Cosserat rod legs. Reaction forces from each leg are applied at the respective attachment points (i.e., the four corners) of the cuboid torso. In this study, a decoupled approach is adopted to reduce computational cost. This decoupled formulation is justified by the large mass disparity between the torso and individual legs and the relatively low angular velocities observed during the considered gaits, which limit the contribution of leg dynamics to the global angular momentum in the tested operating regimes. Under these conditions, the torso mass dominates the inertia contributions of the legs. A modular strategy is employed in which each leg is simulated independently, and the resulting reaction forces are used to update the torso's translational motion. Further, it is assumed that the effect of precession and nutation of the torso is neglected, resulting in,

\begin{equation}
\frac{d}{dt}(I\boldsymbol{\omega}) = I\dot{\boldsymbol{\omega}} + \boldsymbol{\omega} \times (I\boldsymbol{\omega}) \approx I\dot{\boldsymbol{\omega}}
\end{equation}
\noindent
This approximation, made in controllers such as \cite{c24}, is admissible for bodies with small angular velocities For the locomotion scenarios considered in this work, the neglected precession and nutation terms were found to be small relative to the dominant inertial terms, due to the low torso angular velocities during walking and crawling gaits. Thus, the torso's translational and rotational dynamics are governed by:

\begin{equation}
m_c \ddot{\bm{p}} = \sum_{i=1}^{4} \bm{F}_i, \quad
\bm{I}_c \dot{\bm{\omega}} = \sum_{i=1}^{4} (\bm{r}_i \times \bm{F}_i)
\end{equation}
\noindent
where, \(m_c\) is the torso mass, \(\bm{p}\) its center of mass position, and \(\bm{F}_i\) is the reaction force from leg \(i\); \(\bm{I}_c\) is the torso inertia matrix, \(\bm{\omega}\) the angular velocity, and \(\bm{r}_i\) is the vector from the torso's centre of mass to the \(i\)-th leg attachment point.

\section{MODEL PREDICTIVE CONTROL (MPC)}
\begin{figure*}[h!]
    \centering
    \includegraphics[width=1\textwidth]{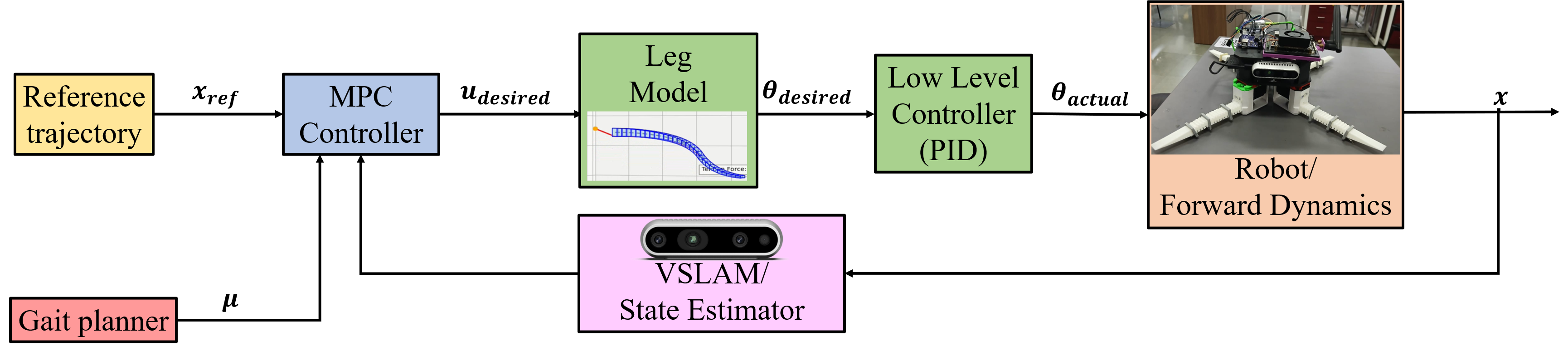} 
    \caption{Model Predictive Control block diagram.}
    \label{fig:mpc_blk_diagram}
    \end{figure*}

MPC is implemented to ensure stable, responsive, and adaptive locomotion for the SLOT robot by predicting and optimizing future states based on current observations. The MPC algorithm integrates a predictive model derived from Cosserat rod dynamics and rigid body torso dynamics to maintain desired trajectories and velocities. The control architecture consists of a hierarchical feedback loop for legged locomotion. A simple block diagram is shown in Fig. \ref{fig:mpc_blk_diagram}. The robot state at timestep $k$ is 
$\bm{x}_k = [\phi,\theta,\psi,\,p_x,p_y,p_z,\,\omega_x,\omega_y,\omega_z,\,v_x,v_y,v_z]^\top$, 
where $(\phi,\theta,\psi)$ are roll, pitch, yaw, $(p_x,p_y,p_z)$ positions, $(\omega_x,\omega_y,\omega_z)$ angular velocities, $(v_x,v_y,v_z)$ linear velocities, and $g_z$ represents gravity. The control input 
$\bm{u} = [F_{x,0},F_{y,0},F_{z,0},\ldots,F_{x,3},F_{y,3},F_{z,3}]^\top$ 
denotes the ground reaction forces in $x,y,z$ for each of the four legs.

A reference trajectory $\bm{x}_{ref}$, together with gait cycle and gait parameter $\bm{\mu}$ from the gait planner, is fed into the MPC controller, which computes the optimal control input $\bm{u}_{desired}$.  The leg model converts this input into joint-level commands $\bm\theta_{\text{desired}}$. These commands are tracked by a low-level PID controller, which compares $\bm\theta_{\text{desired}}$ with the actual joint angles $\bm\theta_{\text{actual}}$ and generates motor torques to minimize the error. The robot, represented by its forward dynamics, executes the motion and produces the actual state $\bm{x}$. This state is estimated through a VSLAM-based state estimator that fuses visual and inertial data. The estimated state $\bm{x}$ is fed back to the MPC controller to close the loop, ensuring accurate trajectory tracking. Thus, the framework integrates high-level planning, mid-level modeling, and low-level control for robust locomotion. 

The average execution times for each component in the control pipeline were measured on the onboard NVIDIA Jetson Xavier platform. The Cosserat rod simulation for each leg required approximately 0.35~ms per integration step, while torso dynamics update required 0.18~ms. The MPC optimization (solved using SCS) required 6.2~ms per update at a control frequency of 30~Hz. The remaining components, including gait scheduling, force-to-angle mapping, and low-level PID control, collectively required less than 0.5~ms per cycle. 

The complete ROS2-based hardware-in-the-loop execution pipeline, detailing perception, state estimation, MPC optimization, and low-level tendon control, is summarized in Algorithm~\ref{alg:ros_mpc}.

    \subsection{Gait Phase Scheduler}
    
    The MPC controller operates based on a predefined walking gait with a 5-second cycle \cite{c23}. Fig. \ref{fig:walk_sim} shows the gait pattern for a forward walk gait. This gait schedule alternates the contraction and relaxation of the front legs while maintaining constant support from the back legs. This alternating activation is encoded as a time-based switching phase and used to modulate friction coefficients and control constraints for the MPC. Similarly, the gait phase scheduler schedules gait for omnidirectional motion as shown in fig. \ref{fig:omni_path}.

    \subsection{Cost Function and Constraints}
    
    The MPC minimizes the following cost function:

\[
J=\sum_{k=0}^{H-1}\Big[
(\boldsymbol{x}_k-\boldsymbol{x}_{\mathrm{ref},k})^{\top}\boldsymbol{Q}\,(\boldsymbol{x}_k-\boldsymbol{x}_{\mathrm{ref},k})
+\boldsymbol{u}_k^{\top}\boldsymbol{R}\,\boldsymbol{u}_k
\Big]
\]

\noindent
where $H=15$ is the prediction horizon, $\boldsymbol{x}_{\mathrm{ref},k}$ denotes the reference trajectory, $\bm{Q}$ is a state weighting matrix prioritizing desired trajectories, and $\bm{R}$ penalizes large control inputs to encourage efficient actuation. The weighting matrices are chosen as:

\[
\resizebox{\columnwidth}{!}{$
\bm{Q}=\tfrac{1}{1000}\,\operatorname{diag}(0.1,0.1,0.1,0,0,100,0.01,0.01,0.01,20,0.1,0,0)
$}
\]

\begin{equation}
  \bm{R} = 10^{-8} \bm{I}_{12}
  \label{R and Q}
\end{equation}

The weighting matrices $\bm{Q}$ and $\bm{R}$ in \eqref{R and Q} were selected to reflect task priorities during locomotion. Higher weights were assigned to the vertical position and forward velocity to ensure stable body support and accurate velocity tracking, while roll and pitch were penalized to limit excessive body tilt. Lower weights were used for less critical states to avoid overly aggressive control actions. The control weighting matrix $\bm{R}$ was chosen to balance control authority and smooth actuation. All weights were tuned iteratively in simulation and kept fixed across all experiments.

\noindent    
The control inputs are constrained by:
    \[
    f_{z,i} \leq 6.0 , \sum_{i=1}^{4} f_{z,i} = mg, \text{ and } |f_{x,i}| \leq \mu_i f_{z,i}, |f_{y,i}| \leq \mu_i f_{z,i}
    \]
    where $f_{z,i}$ is the normal force exerted by leg $i$, $f_{\text{max}} = 6.0$ N is the maximum allowable force per leg, and $\mu_i$ is the friction coefficient at leg $i$, varying with gait phase (e.g., 0.6 or 0.2 for front legs, 0.1 for back legs in case of Walk gait).
    
\subsection{Solver and Real-Time Loop}

All simulations and control implementations in this work were carried out using \texttt{Python}. The Cosserat rod-based leg dynamics and the rigid-body torso dynamics were implemented from scratch using the \texttt{NumPy} and \texttt{SciPy} scientific computing libraries. Visualization and post-processing were performed using \texttt{Matplotlib}.

The Model Predictive Control (MPC) problem was formulated as a convex quadratic program and solved using the Splitting Conic Solver (SCS) via the \texttt{CVXPY} optimization framework. The MPC operates at a control frequency of 30~Hz with a prediction horizon of 15 steps (approximately 0.495~s). At each interval, optimal leg forces are computed to maintain desired forward velocity ($v_x = 0.08$ m/s) and vertical body support ($p_z = 0.04$ m). In addition, the cost function penalizes deviations in roll and pitch angles, encouraging minimal body tilt during locomotion. This helps ensure stable walking and consistent foot-ground contact over time. The MPC runs in batches (20 steps) followed by forward dynamics (5 steps) for real-time feasibility. The batches of MPC optimizations are followed by short forward-dynamics rollouts for analysis efficiency; during execution, the MPC is solved at every control timestep and no MPC updates are skipped. All experiments were executed on an onboard NVIDIA Jetson Xavier platform, matching the computational constraints of the physical robot. This unified software pipeline ensures consistency between simulation and experimental deployment.
    
    \subsection{Leg Model and Low-Level Control}
    
    The leg model converts MPC-generated ground reaction force commands into tendon actuation angles. Specifically, each leg’s desired normal force $f_{z,i}$ is mapped to a servo pulling angle $\theta_i$ using a closed-form relation derived from the Cosserat rod simulation of the leg. By simulating a series of single-leg Cosserat rod deformations under varying tendon angles, an empirical fit was obtained to relate the resulting vertical ground reaction force to the pulling angle:
    \begin{equation}
    \bm\theta_i = \frac{\textbf{$f_{z,i}$} + 6.91}{0.107}
    \end{equation}
    Dynamic effects such as tendon compliance, cable friction, TPU hysteresis, and motor dynamics are not explicitly modeled in this mapping and are instead handled implicitly by feedback control and the receding-horizon MPC framework.
    The low-level PID controller compares desired joint angles $\theta_{\text{desired}}$ with actual joint angles $\theta_{\text{actual}}$ and computes corrective motor commands. This ensures accurate tracking and smooth execution of the robot’s leg movements.

For clarity and reproducibility, Algorithm~\ref{alg:ros_mpc} outlines the full closed-loop execution sequence implemented on the physical robot.

\begin{algorithm}[H]
\caption{ROS2-Based Hardware-in-the-Loop MPC Locomotion Pipeline for SLOT}
\label{alg:ros_mpc}
\small
\begin{algorithmic}[1]

\STATE \textbf{System Initialization:}
\STATE Initialize ROS2 middleware and time synchronization
\STATE Load robot geometric, inertial, and Cosserat rod leg parameters
\STATE Initialize onboard computation for perception and control
\STATE Configure publishers and subscribers for all control nodes

\vspace{2pt}
\STATE \textbf{Perception and Localization:}
\STATE Start visual--inertial SLAM using RGB-D and IMU inputs
\STATE Publish robot odometry and pose estimates on \texttt{/odom}
\STATE Broadcast frame transformations via \texttt{/tf}

\vspace{2pt}
\STATE \textbf{State Estimation Node:}
\STATE Subscribe to \texttt{/odom} and inertial measurements
\STATE Extract position, orientation, linear and angular velocities
\STATE Construct MPC state vector
\[
\bm{x}_k =
[\phi,\theta,\psi,\,p_x,p_y,p_z,\,\omega_x,\omega_y,\omega_z,\,v_x,v_y,v_z]
\]
\STATE Publish estimated state on \texttt{/mpc\_state}

\vspace{2pt}
\STATE \textbf{Controller Initialization:}
\STATE Initialize gait scheduler and phase manager
\STATE Initialize MPC controller with horizon $H$, timestep $\Delta t$, and cost weights
\STATE Initialize low-level tendon motor controllers

\vspace{4pt}
\WHILE{robot is operational}

    \STATE \textbf{State Update:}
    \STATE Receive latest robot state $\bm{x}_k$ from \texttt{/mpc\_state}

    \vspace{2pt}
    \STATE \textbf{Gait and Reference Generation:}
    \STATE Update gait phase (walk, crawl, omnidirectional)
    \STATE Determine stance and swing legs
    \STATE Generate desired body height and velocity references
    \STATE Form reference trajectory $\bm{x}_{\mathrm{ref},k}$

    \vspace{2pt}
    \STATE \textbf{MPC Optimization:}
    \STATE Linearize decoupled torso dynamics about $\bm{x}_k$
    \STATE Formulate convex MPC using predicted ground reaction forces
    \STATE Enforce force balance, friction cone, and contact constraints
    \STATE Solve for optimal control input
    \[
    \bm{u}_k^\star =
    [f_{x,1},f_{y,1},f_{z,1},\dots,f_{x,4},f_{y,4},f_{z,4}]
    \]
    \IF{optimization is infeasible}
        \STATE Apply last feasible control input
    \ENDIF

    \vspace{2pt}
    \STATE \textbf{Force-to-Actuation Mapping:}
    \FOR{$i = 1$ to $4$}
        \STATE Convert desired normal force $f_{z,i}$ to tendon angle $\theta_i$
        \STATE Use precomputed Cosserat rod–based force--angle relation
    \ENDFOR
    \STATE Publish desired tendon angles on \texttt{/desired\_joint\_angles}

    \vspace{2pt}
    \STATE \textbf{Low-Level Motor Control:}
    \STATE Subscribe to \texttt{/desired\_joint\_angles}
    \STATE Track tendon angles using PID control
    \STATE Compensate actuator and tendon dynamics via feedback

    \vspace{2pt}
    \STATE \textbf{Execution and Feedback:}
    \STATE Apply motor commands to the robot
    \STATE Execute compliant leg motion
    \STATE Advance control loop by $\Delta t$

\ENDWHILE

\end{algorithmic}
\end{algorithm}

\subsection{Stability Analysis}

 Analyzing the long-term behavior of the MPC solution for the quadruped robot is crucial to ensure recursive feasibility and robustness. Recursive feasibility implies that the MPC problem remains solvable at each time step, satisfying all constraints (e.g., force limits, friction coefficients, and total force balance) throughout the simulation. To visually assess this property, a Dynamic Constraint Map (DCM) is employed, providing a comprehensive representation of constraint satisfaction across time and variables as a heatmap.

Stability is also evaluated through a series of perturbation tests conducted over a 25 s simulation horizon, comprising 15 MPC optimization steps followed by 5 forward dynamics rollouts per cycle. These tests assess the MPC's robustness across seven distinct scenarios: Baseline (initialized at the reference state with $p_z = 0.04$ m and $v_x = 0.08$ m/s), Roll (initial roll angle of 0.1 rad), Pitch (initial pitch angle of 0.1 rad), Height ($p_z = 0.01$ m), Velocity ($v_x = 0.05$ m/s), Combined (initial roll and pitch angles of 0.05 rad each, $p_z = 0.02$ m, $v_x = 0.03$ m/s), and Noise (Gaussian-distributed perturbations with standard deviation $\sigma = 0.001$ applied to all states, followed by clipping to prevent divergence). Key performance metrics encompass the maximum, mean, and final costs; settling time (defined as the first instant when the cost remains below 0.01 persistently); and the number of infeasible optimization steps.

\section{EXPERIMENTAL VALIDATION AND RESULTS}

Comparative analyses between simulated trajectories and experimental measurements from a physical prototype, equipped with ArUco markers, were conducted to validate the proposed Cosserat rod-based tendon-actuated leg model. Validation included single-leg tests, crawling, forward walking and omnidirectional gaits.

\subsection{Model Validation}
Leveraging the decoupled Cosserat rod model developed in Section IV, this section outlines a validation pipeline to assess its accuracy in predicting SLOT’s whole-body locomotion. Experimental ground-truth measurements were obtained using an Intel RealSense D435 RGB-D camera operating at 30~Hz, with depth accuracy of approximately $\pm2\%$ at a 1~m range. ArUco marker detection and pose estimation were performed using OpenCV-based methods. The validation pipeline consists of the following steps: 1) Data Collection: ArUco markers are placed on the physical robot at key locations (CoM and leg segments (base, one-third length, two-third length, and tip)) to capture 3D motion trajectories using an external camera system; 2) Simulation Execution: The SLOT simulation is executed using predefined gaits (crawl, walk) with corresponding tendon schedules; 3) Time Alignment: Experimental data is aligned with simulated time vectors via interpolation to ensure synchronized comparisons; 4) Metric Computation: Error metrics (RMSE, MAE, NRMSE) are computed over key points on the body; 5) Visualization and Logging: Plots comparing simulation and experiment are generated for each gait, with detailed annotations to highlight agreement and deviations.

This validation framework ensures repeatability, transparency, and quantitative rigor in evaluating the Cosserat rod model's predictive performance for whole-body locomotion in soft quadrupeds.

\subsection{Single-Leg Validation}
\begin{figure*}[h!]
    \centering
    \includegraphics[width=0.8\textwidth]{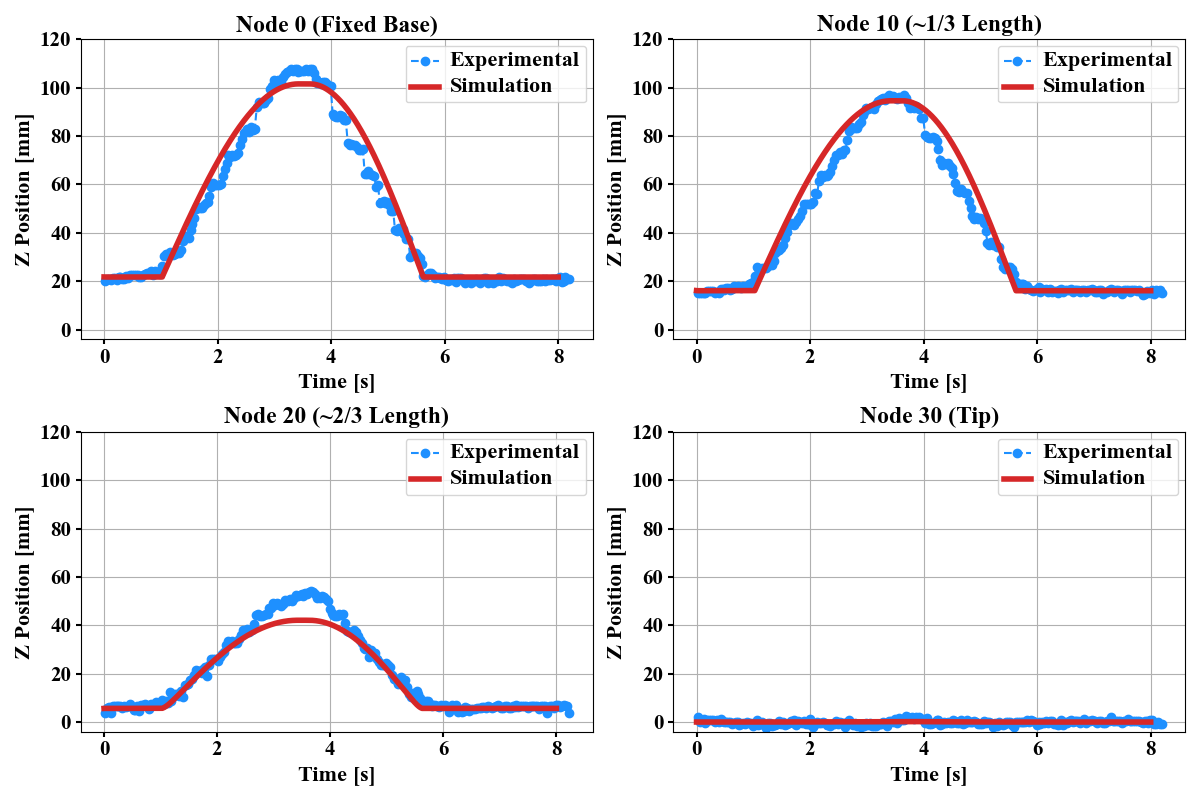}
    \caption{Comparison of experimental and simulated vertical position $z(t)$ at four locations along the single leg—Node~0 (base), Node~10 (one-third length), Node~20 (two-thirds length), and Node~30 (tip).}
    \label{fig:z_position_comparison_single_leg}
\end{figure*}

A single-leg simulation was conducted to validate the core Cosserat rod model. Four representative points - fixed base, 1/3 leg length, 2/3 leg length, and tip - were tracked experimentally and compared with simulation outputs as shown in Fig. \ref{fig:z_position_comparison_single_leg}. The model showed strong agreement in vertical deformation patterns under tendon actuation.

\begin{table*}[h!]
\centering
\caption{Single-Leg Validation Error Metrics}
\begin{tabular}{|l|c|c|c|c|}
\hline
\textbf{Metric} & \textbf{Fixed Base} & \textbf{1/3 Length} & \textbf{2/3 Length} & \textbf{Tip} \\ \hline
RMSE (mm)        & 5.20  & 5.72  & 4.00  & 0.89 \\ 
MAE (mm)         & 3.85  & 3.90  & 2.60  & 0.71 \\ 
NRMSE (\%)       & 5.87   & 6.94   & 7.88   & 19.22 \\ 
Avg Error (mm)   & 2.23  & 2.87  & 2.32 & 0.15 \\ 
Error (\%)       & 2.51   & 3.47   & 4.56   & 3.21  \\ 
Accuracy (\%)    & 97.49  & 96.53  & 95.44  & 96.79 \\ \hline
\end{tabular}
\label{tab:single_leg_metrics}
\end{table*}

As it can be seen in Table~\ref{tab:single_leg_metrics}, the RMSE and MAE values are larger near the fixed base and one-third leg length, indicating greater absolute deviations at these segments due to their higher rigidity and resulting stress concentrations under tendon loading. The highest NRMSE observed at the leg tip (19.22\%) arises from normalization over a small total displacement at this point, causing even minor absolute deviations to appear significant in relative terms and not reflecting large absolute positional errors. Overall accuracy remains high (above 95\%) across all points, demonstrating that despite minor localized deviations, the Cosserat rod model robustly predicts overall leg deformation.

These insights help pinpoint specific areas of the model requiring fine-tuning and also affirm the model’s effectiveness in capturing complex deformation behaviors accurately.

\subsection{Forward Walking Gait Validation}
\begin{figure*}[t]
    \centering
    \includegraphics[width=\textwidth]{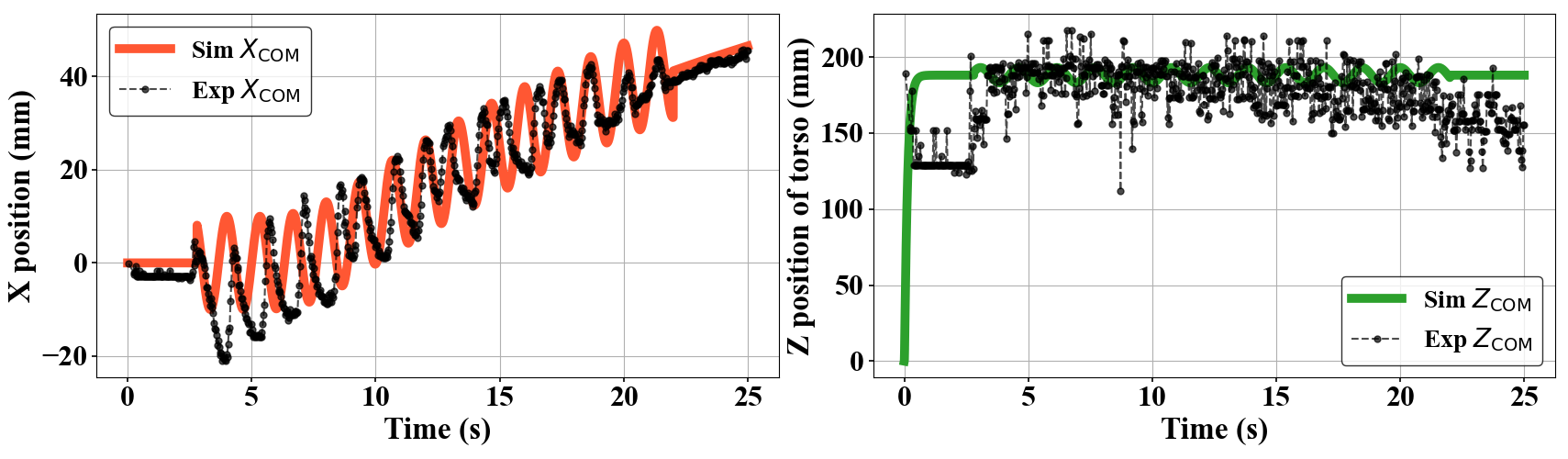}
    \caption{Comparison of simulated and experimental center-of-mass (CoM) trajectories for the forward walking gait in the $x$ and $z$ directions.}
    \label{fig:Forward_walk_com_comparison}
\end{figure*}

\begin{figure*}[t]
    \centering
    \includegraphics[trim=0 20 0 0,clip,width=\textwidth]{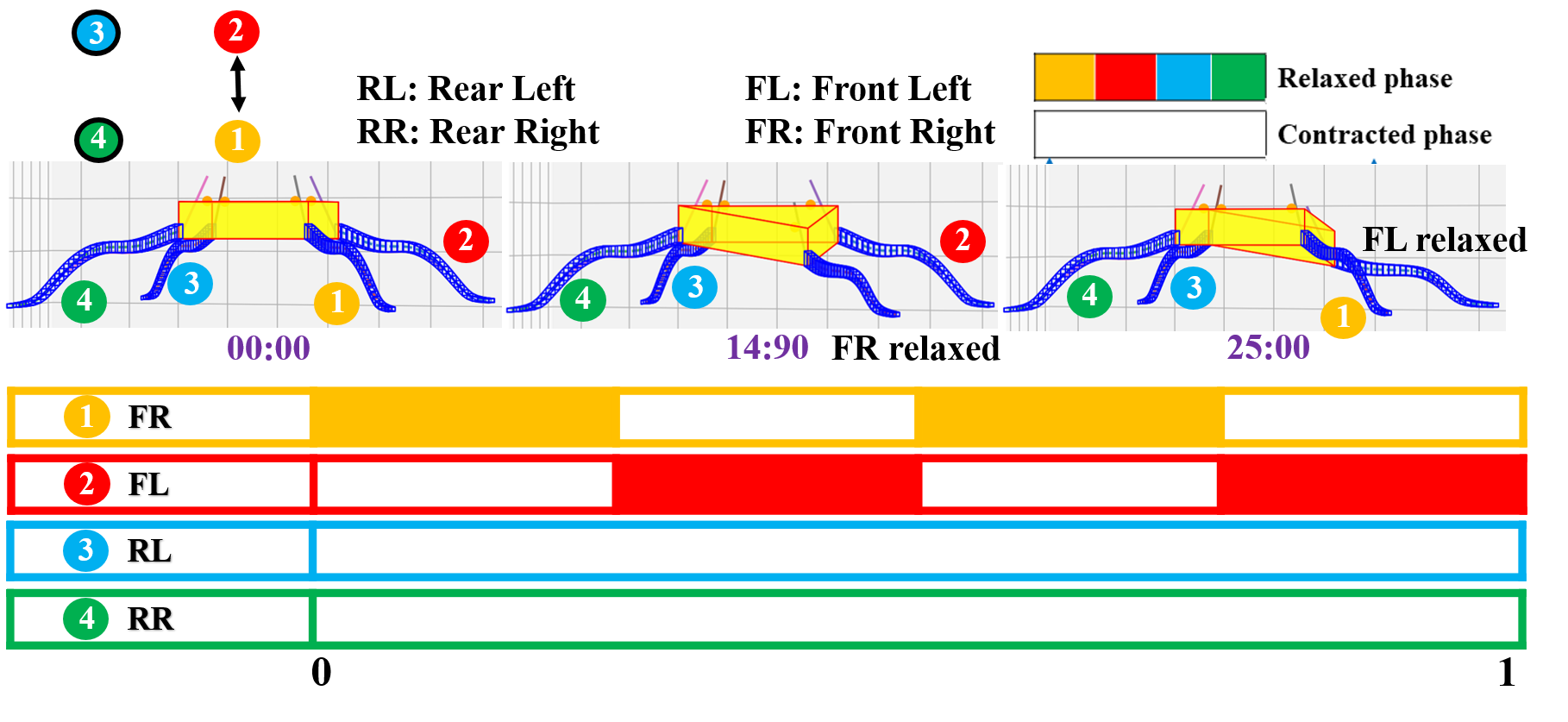}
    \caption{SLOT walking in a straight line in the simulation environment.}
    \label{fig:walk_sim}
\end{figure*}

As shown in Fig. \ref{fig:Forward_walk_com_comparison}, using a two-phase gait profile, the forward walking simulation matched experimental trends in forward propulsion and vertical body support. Fig. \ref{fig:walk_sim} shows the timelapse in simulation. The
colored circles near the legs denote contraction of the respective leg.

Overall, these results confirm that the Cosserat rod model offers physically realistic predictions of soft quadruped motion across diverse gaits.
\subsection{Crawling Gait Validation}
\begin{figure*}[h!]
    \centering
    \includegraphics[width=1\textwidth]{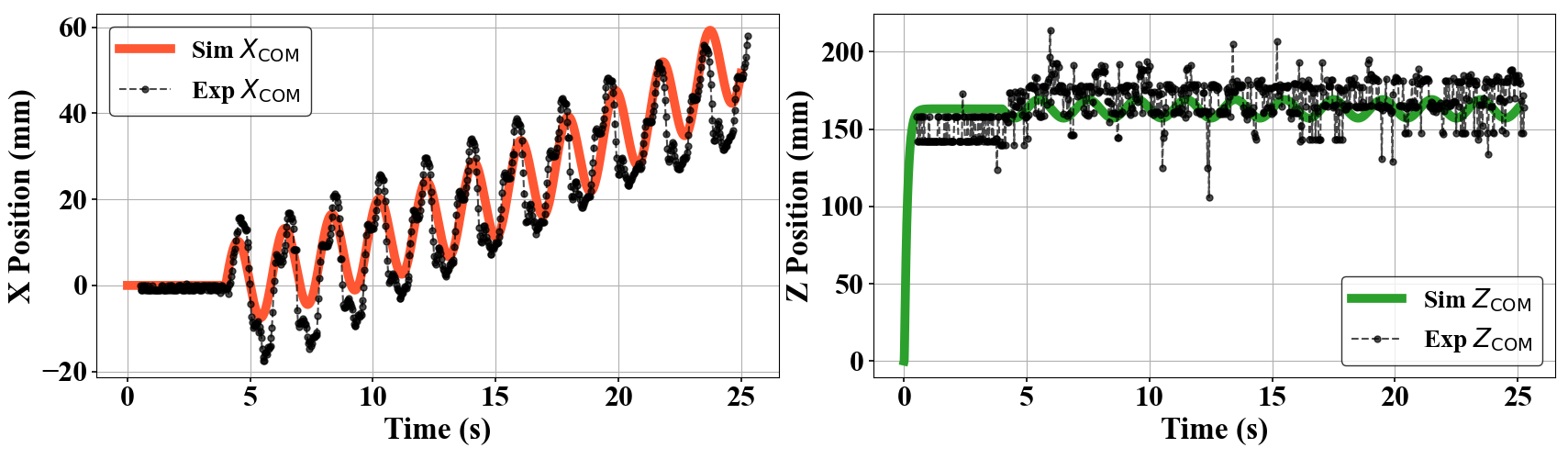}
    \caption{Comparison of simulated versus experimental center-of-mass (CoM) trajectories for the crawl gait.}
    \label{fig:crawl_COM_xyz_comparison}
\end{figure*}

The crawling gait simulation followed a four-phase activation profile. The resulting center of mass (CoM) trajectories in X and Z directions were compared with experimental data as shown in Fig. \ref{fig:crawl_COM_xyz_comparison}. The model accurately reproduced the crawling behavior, with consistent ground contact and compliant vertical modulation.

This confirms the model’s versatility across gaits with an RMSE shown in Table \ref{tab:whole_body_metrics} for Forward Walk and Crawl gait respectively.

\begin{table}[t]
\centering
\caption{Quantitative Validation Metrics for Whole-Body Locomotion}
\label{tab:whole_body_metrics}
\resizebox{\columnwidth}{!}{%
\begin{tabular}{|l|c|c|}
\hline
\textbf{Gait} & \textbf{RMSE in $x_{\text{CoM}}$ (mm)} & \textbf{RMSE in $z_{\text{CoM}}$ (mm)} \\ \hline
Forward Walk  & 4.0 & 4.8 \\ 
Crawl         & 5.1 & 4.6 \\ \hline
\end{tabular}
}
\end{table}

\subsection{MPC-Driven Locomotion: Feasibility and Stability}
\begin{figure*}[t]
    \centering

    \begin{subfigure}{0.48\textwidth}
        \centering
        \includegraphics[width=\linewidth]{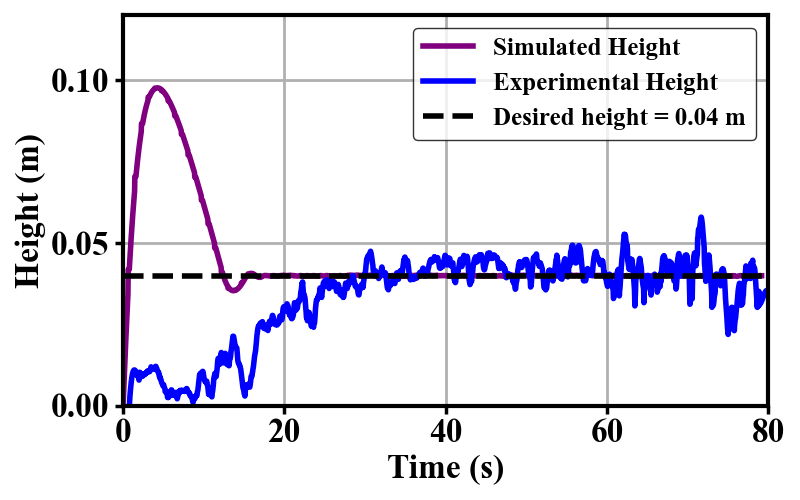}
        \caption{Height tracking}
        \label{fig:mpc_pz}
    \end{subfigure}
    \hfill
    \begin{subfigure}{0.48\textwidth}
        \centering
        \includegraphics[width=\linewidth]{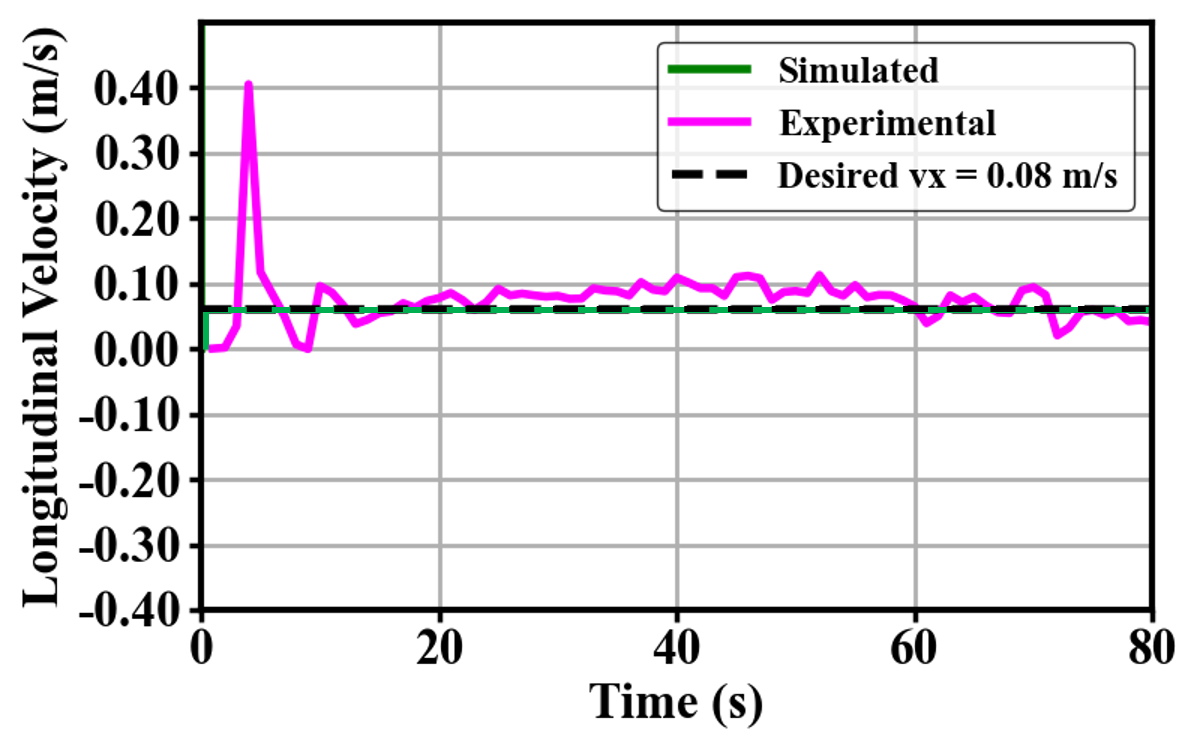}
        \caption{Longitudinal velocity}
        \label{fig:mpc_vx}
    \end{subfigure}

    \begin{subfigure}{0.48\textwidth}
        \centering
        \includegraphics[width=\linewidth]{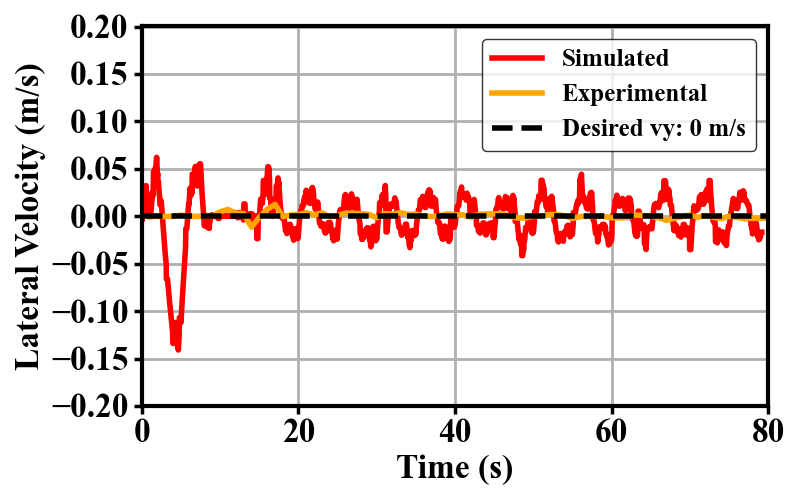}
        \caption{Lateral velocity}
        \label{fig:mpc_vy}
    \end{subfigure}
    \hfill
    \begin{subfigure}{0.48\textwidth}
        \centering
        \includegraphics[width=\linewidth]{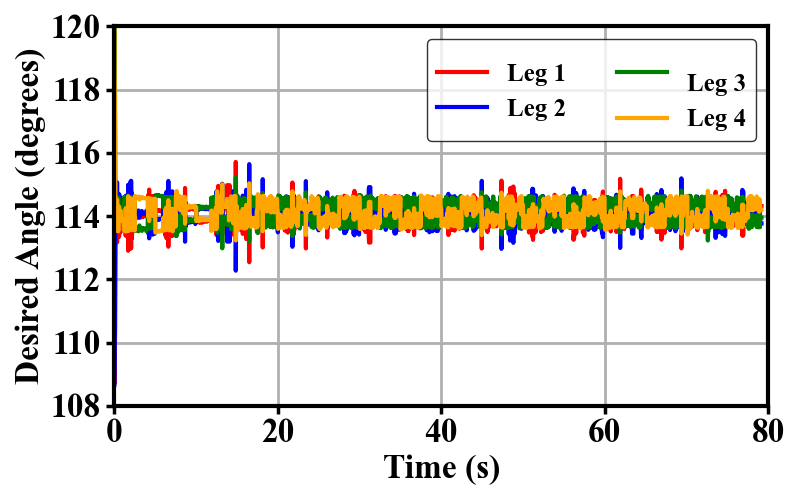}
        \caption{Desired joint angles}
        \label{fig:mpc_des_angles}
    \end{subfigure}

    \caption{Experimental and simulated results of quadruped MPC for the walking gait. Leg 1: FL, Leg 2: RL, Leg 3: RR, Leg 4: RL.}
    \label{fig:mpc_results}
\end{figure*}

Fig. \ref{fig:mpc_results} shows SLOT reaches a desired height of 0.04 metres, achieves forward velocity of 0.08 m/s while maintaining near-zero mean lateral velocity over each gait cycle for the walk gait.

\begin{figure*}[t]
    \centering
    \includegraphics[width=\textwidth]{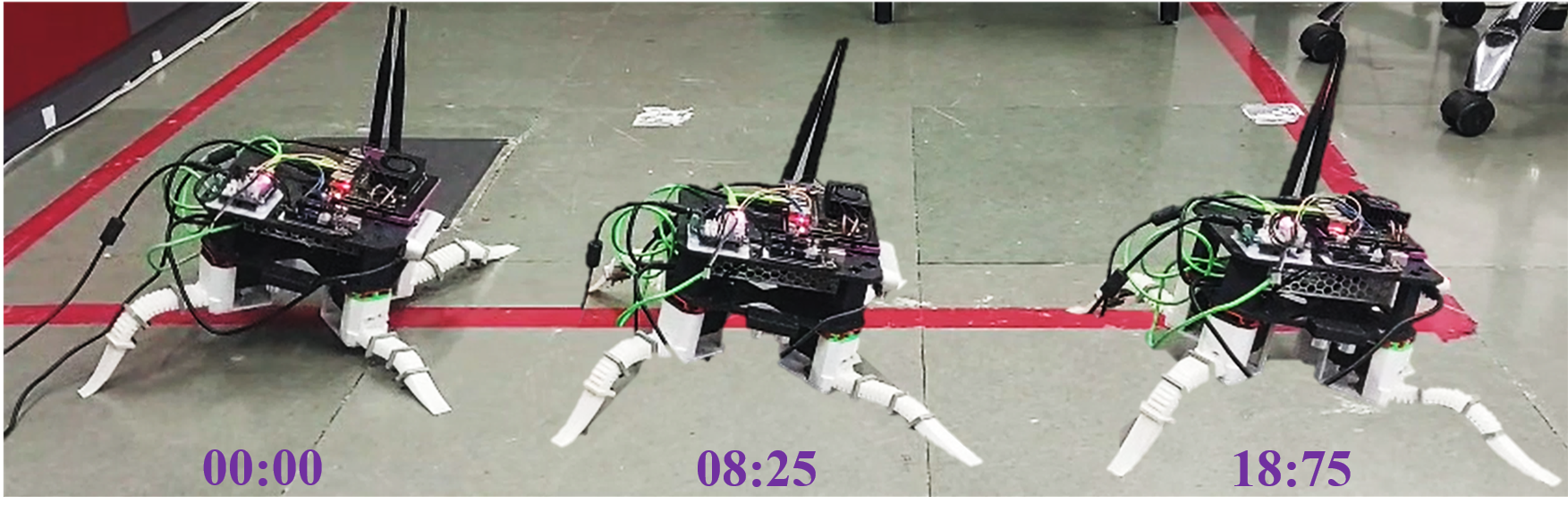}
    \caption{SLOT hardware robot walking in a straight line under MPC control, demonstrating stable forward locomotion.}
    \label{fig:walk_timelapse}
\end{figure*}

Additionally, Fig. \ref{fig:walk_timelapse} provides a time-lapse sequence of the robot walking in a straight line under MPC control, demonstrating stable forward locomotion and visually confirming the controller’s effectiveness in maintaining gait consistency.

Over a forward displacement of approximately $0.8$~m, the maximum lateral deviation was $3.2$~cm, corresponding to an average lateral RMSE of $1.4$~cm. This drift is primarily attributed to unmodeled ground friction asymmetries and hardware tolerances, and does not affect local stability or vertical support, which are the primary objectives of the MPC formulation.

The experimental ground-truth measurements were obtained using an Intel RealSense D435 RGB-D camera. According to the manufacturer’s specifications and prior benchmarking studies, the depth accuracy of the D435 is approximately $\pm 2\%$ at a range of 1~m, with sub-millimeter repeatability under indoor lighting conditions \cite{c28}. In our experiments, the robot operated within a distance range of 0.6--1.0~m from the camera, resulting in an expected depth uncertainty below 2~mm. This measurement noise is significantly smaller than the observed center-of-mass displacements and therefore does not affect the reported RMSE values or the qualitative conclusions drawn from the experimental validation.

\begin{figure*}[t]
    \centering

    \begin{subfigure}{0.48\textwidth}
        \centering
        \includegraphics[width=\linewidth]{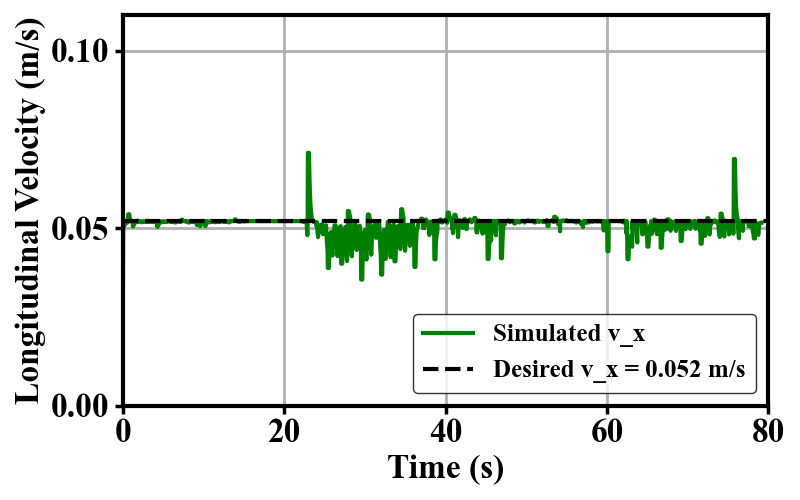}
        \caption{Longitudinal velocity}
        \label{fig:omni_longitudinal}
    \end{subfigure}
    \hfill
    \begin{subfigure}{0.48\textwidth}
        \centering
        \includegraphics[width=\linewidth]{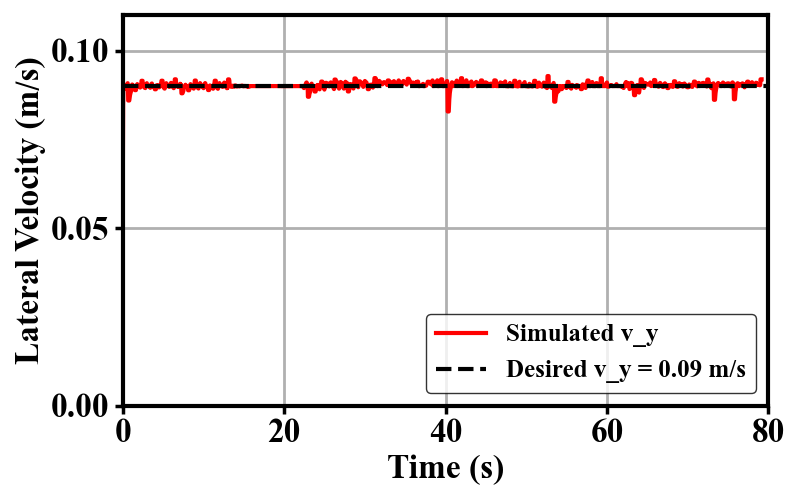}
        \caption{Lateral velocity}
        \label{fig:omni_lateral}
    \end{subfigure}

    \begin{subfigure}{\textwidth}
        \centering
        \includegraphics[width=0.85\linewidth]{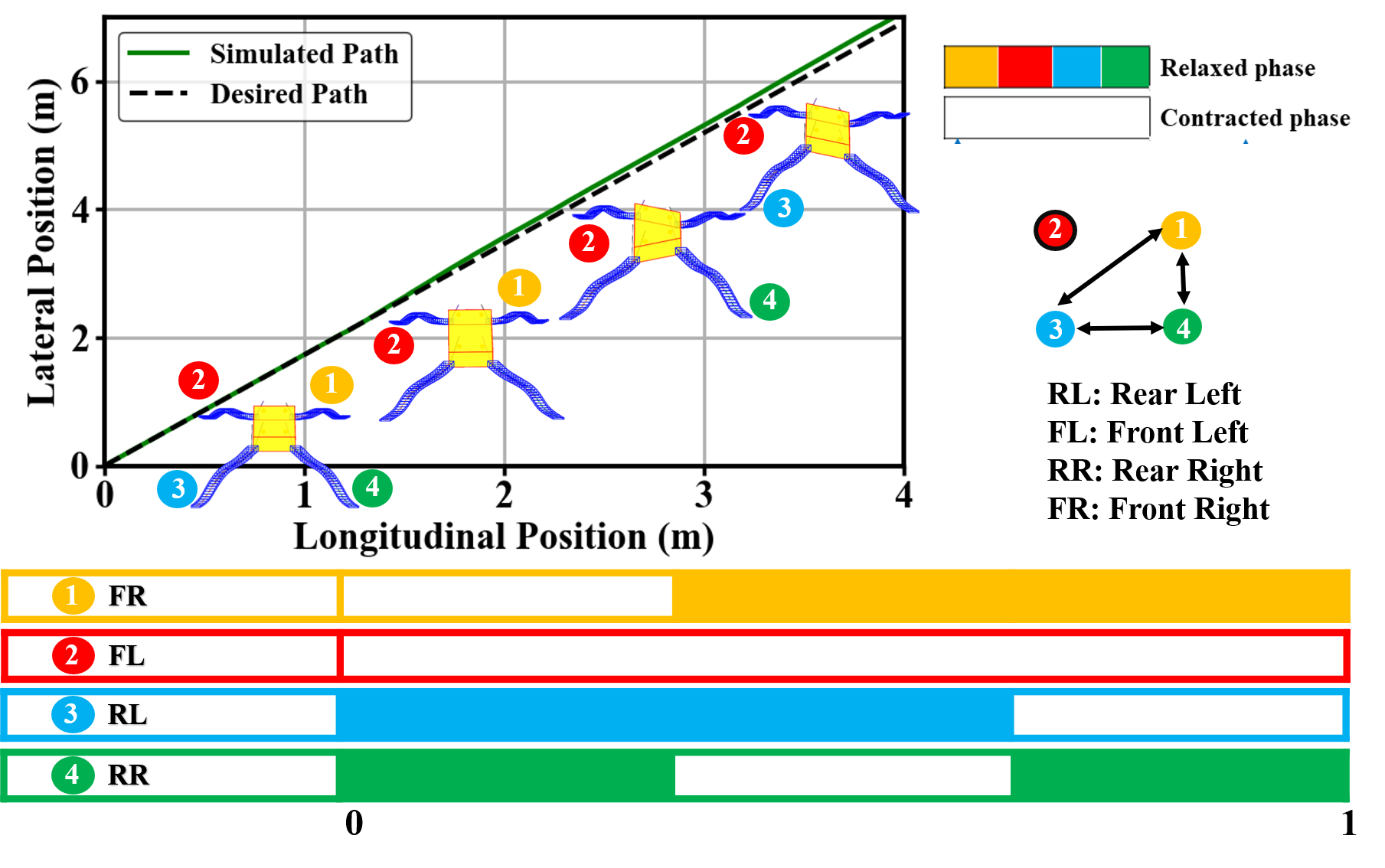}
        \caption{Desired versus simulated path with desired gait sequence. Colored circles near the legs indicate leg contraction.}
        \label{fig:omni_path}
    \end{subfigure}

    \caption{Simulation results for the $60^\circ$ omnidirectional gait under MPC.}
    \label{fig:overall}
\end{figure*}

Figure~\ref{fig:overall} summarizes the $60^\circ$ omnidirectional gait regulated by the MPC. In Fig. \ref{fig:omni_longitudinal}, the longitudinal velocity $v_x$ closely follows the commanded constant $v_x^\star = 0.052~\mathrm{m/s}$, with brief transients at phase switches. Fig. \ref{fig:omni_lateral} shows similar behavior for the lateral velocity $v_y$, tracking $v_y^\star = 0.09~\mathrm{m/s}$; the small ripple visible in both traces is attributable to contact transitions and periodic leg actuation. Fig. \ref{fig:omni_lateral} compares the desired diagonal path with the simulated body trajectory: the net motion aligns with the commanded direction since $v_y^\star / v_x^\star \approx 0.09/0.052 \approx 1.73 \simeq \tan 60^\circ$ with an RMSE of 3.5 cm. The simulated robot's snapshots illustrate progression along the path, while numbered colored circles indicate which leg is contracting at each step. The gait plot beneath encodes the stance/relaxation timing for all four legs over one stride and matches the shown sequence.

\begin{figure*}[htbp]
\centering
\includegraphics[width=0.8\textwidth]{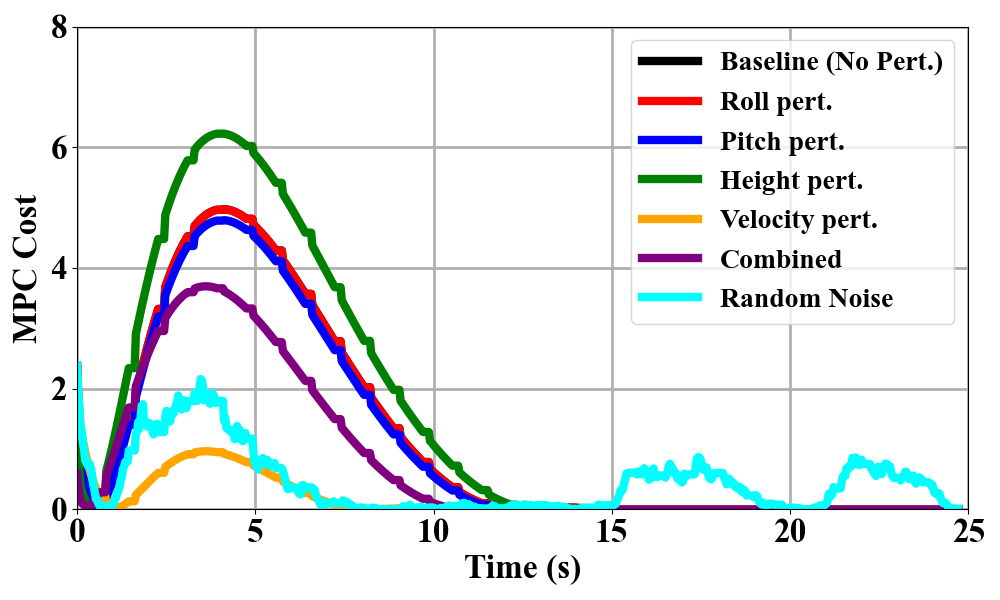}
\caption{Combined MPC cost trajectories under perturbations for Walk gait. Baseline and Roll overlap; Noise exhibits non-settling oscillations.}
\label{fig:perturbations}
\end{figure*}

Table~\ref{tab:mpc_metrics} summarizes the MPC stability metrics using the Cosserat-integrated  for walk gait. All deterministic perturbations (Baseline, Roll, Pitch, Height, Velocity, Combined) achieve zero final cost and settle within 14.553 s, confirming asymptotic stability and recursive feasibility (0 infeasible steps across cases). The Baseline, starting at the reference trajectory, provides the nominal benchmark. Height perturbation incurs the highest maximum cost (6.2248) and integral (38.0922), driven by the $z$ weight. Velocity settles fastest (10.098 s) with the lowest mean cost (0.1720), reflecting efficient correction of $v_x$ error. The Combined case shows moderate performance (max cost 3.6959, settling 12.606 s), demonstrating robustness to multi-axis disturbances. The Noise case ($\sigma = 0.001$) does not settle, with a final cost of 0.4943 due to persistent oscillations, despite a low mean cost (0.2212). Unlike deterministic perturbations (e.g., initial roll=0.1 rad), this is an ongoing disturbance, preventing the system from converging to a stable equilibrium. This behavior is expected, as the applied noise constitutes a persistent disturbance rather than a bounded initial perturbation. Figure~\ref{fig:perturbations} illustrates the combined MPC cost trajectories. Baseline and Roll curves overlap initially (max costs 4.9739 and 4.9719), indicating negligible impact from small roll error.

\begin{table*}[htbp]
\centering
\caption{Stability Metrics for Perturbations (Walk gait)}
\label{tab:mpc_metrics}
\begin{tabular}{lccccc}
\hline
Perturbation & Max Cost & Mean Cost & Final Cost & Settling Time (s) \\
\hline
Baseline     & 4.9739   & 1.1665    & 0.0000     & 14.553          \\
Roll         & 4.9719   & 1.1659    & 0.0000     & 14.553          \\
Pitch        & 4.7920   & 1.1120    & 0.0000     & 13.695          \\
Height       & 6.2248   & 1.5397    & 0.0000     & 14.553          \\
Velocity     & 2.3827   & 0.1720    & 0.0000     & 10.098           \\
Combined     & 3.6959   & 0.7873    & 0.0000     & 12.606          \\
Noise        & 2.3828   & 0.2212    & 0.4943     & Did not settle           \\
\hline
\end{tabular}
\end{table*}

\begin{figure*}[htbp]
    \centering
    \begin{subfigure}[b]{0.45\textwidth}
        \centering
        \includegraphics[trim=0 0 130 0,clip,width=\linewidth]{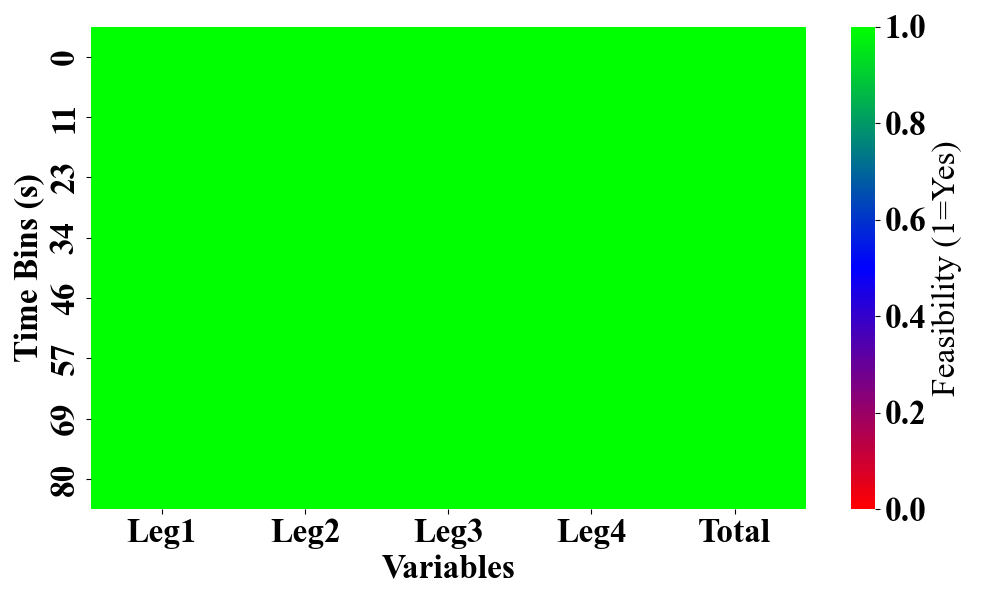}
        \caption{Walk gait}
        \label{fig:dcm_walk}
    \end{subfigure}
    \hfill
    \begin{subfigure}[b]{0.53\textwidth}
        \centering
        \includegraphics[trim=0 0 30 0,clip,width=\linewidth]{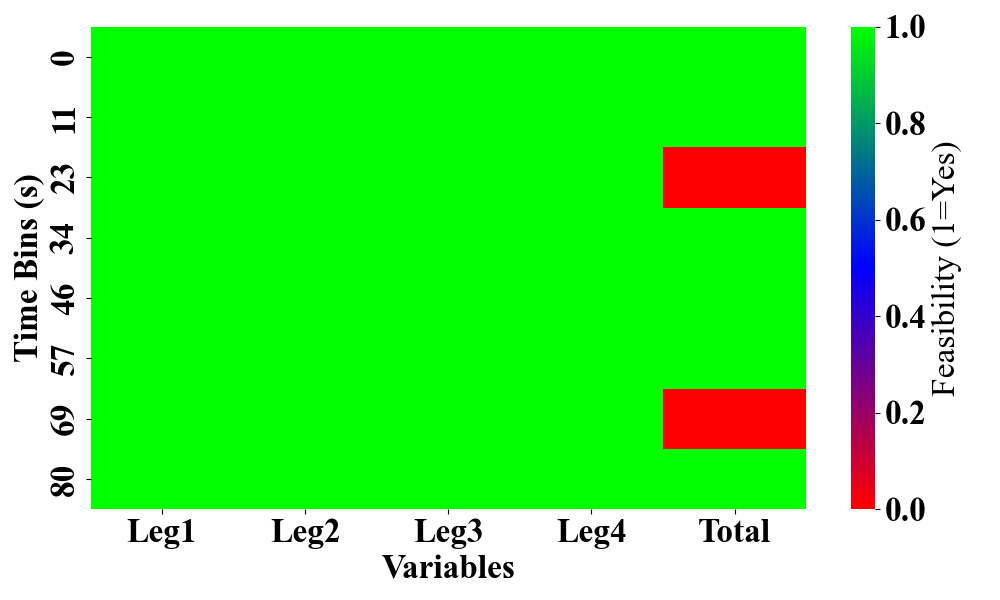}
        \caption{Omnidirectional gait}
        \label{fig:dcm_omni}
    \end{subfigure}
    
    \caption{Dynamic Constraint Map (DCM) illustrating the feasibility status of the MPC solution.}
    \label{fig:feasibility_subplots}
\end{figure*}

The Dynamic Constraint Map, depicted in Figure~\ref{fig:feasibility_subplots}, illustrates the feasibility status for each of the four legs and the total system over the simulation period. In particular, Figure \ref{fig:dcm_walk} shows the walk gait, while Figure \ref{fig:dcm_omni} presents the omnidirectional gait. The map uses a color gradient where red indicates infeasibility, whereas green indicates feasibility. The map predominantly displays a uniform green color for walk gait, indicating consistent feasibility across all time steps and variables. For omnidirectional gait, the presence of red areas in the "Total" column, suggests that the overall system feasibility was lost during those intervals. This could affect the robot's stability or ability to follow the desired trajectory. The fact that Legs 1-4 show green indicates that individual leg constraints might still be feasible, but the combined system constraints were violated at 22.374 and 61.875 seconds. When infeasibility is detected during simulation, the controller reverts to the most recent feasible control input, preventing divergence and ensuring continued execution.

\section{DISCUSSION AND CONCLUSION}

This work presents a detailed modeling and validation framework for SLOT, a soft-legged omnidirectional tetrapod robot actuated via tendon-driven flexible legs. By leveraging Cosserat rod theory, we accurately model the continuous deformation behavior of soft TPU legs under real-time actuation and ground interaction. The use of a decoupled torso-leg formulation offers significant computational efficiency. This modular architecture is especially well-suited for MPC for soft mobile robots, where scalability and latency are key challenges.

Experimental validation across single-leg and whole-body locomotion scenarios confirms the physical fidelity of the proposed model. The single-leg validation achieved over 95\% accuracy in Z-position tracking across key nodes, while crawling and walking gait comparisons showed strong alignment in CoM trajectories with 4-5 mm RMSE. These results demonstrate the robustness of the modeling approach and its ability to generalize across different motion patterns.

These results confirm that the MPC achieves asymptotic stability with fast transient response and sustained equilibrium. The absence of infeasibility or cost divergence further validates the controller design for the given operating conditions.

Overall, this study establishes a foundational framework for physics-informed modeling of soft quadrupeds. Future work will focus on real-time adaptation of the model for unstructured terrain. The SLOT platform, with its low-cost open-source design, serves as a testbed for advancing control and learning in mobile soft robotics.

\section{FUTURE WORK}
Building on this foundation, future work will focus on developing a coupled torso-leg dynamic model that retains the physical accuracy of the current framework while improving computational efficiency through reduced-order formulations and parallelization strategies. This will allow tighter integration between torso motion and leg compliance, especially during unstructured terrain interactions. Additionally, we plan to incorporate model-based reinforcement learning (MBRL) to learn adaptive control policies directly from simulation, leveraging the physics-based model as a differentiable prior. This hybrid approach will allow data-efficient policy learning while preserving physical plausibility.

\section{ACKNOWLEDGEMENT}
The authors thank the Government of India for supporting the Saumya Karan's and Suraj Borate's research through the Prime Minister’s Research Fellowship (PMRF) Scheme.  Finally the authors thank IIT Gandhinagar and its donors for their unwavering support.

\section{FUNDING}
The authors received financial support from the Prime Minister’s Research Fellowship (PMRF), Government of India (Project File No. MIS/IITGN/PMRF/ME/MV/2023-24/053).

\end{document}